\documentclass[letterpaper, 10 pt, conference]{ieeeconf}  
\usepackage[utf8]{inputenc}

\IEEEoverridecommandlockouts                              

\usepackage{float}
\usepackage{amsmath} 
\usepackage{amssymb}  
\usepackage{booktabs} 
\usepackage{algorithm} 
\usepackage{algorithmic} 
\usepackage{multirow} 
\usepackage[T1]{fontenc}
\usepackage[utf8]{inputenc}

\usepackage[most]{tcolorbox}
\usepackage{xcolor}
\usepackage{bbold}
\usepackage{url}
\usepackage{hyperref}
\title{\LARGE \bf
Generalizable Dense Reward for Long-Horizon Robotic Tasks
}

\author{
Silong Yong$^{1}$, Stephen Sheng$^{2}$, Carl Qi$^{3}$, Xiaojie Wang$^{2}$\\
Evan Sheehan$^{2}$, Anurag Shivaprasad$^{2}$, Yaqi Xie$^{1}$\\
Katia Sycara$^{1}$, Yesh Dattatreya$^{2}$\\[1ex]
$^{1}$Carnegie Mellon University \quad
$^{2}$Amazon Robotics \quad
$^{3}$UT Austin\\
}

\begin{document}

\maketitle
\thispagestyle{empty}
\pagestyle{empty}

\begin{abstract}Existing robotic foundation policies are trained primarily via large-scale imitation learning.
While such models demonstrate strong capabilities, they often struggle with long-horizon tasks due to distribution shift and error accumulation. While reinforcement learning (RL) can finetune these models, it cannot work well across diverse tasks without manual reward engineering. We propose VLLR, a dense reward framework combining (1) an extrinsic reward from Large Language Models (LLMs) and Vision-Language Models (VLMs) for task progress recognition, and (2) an instrinsic reward based on policy self-certainty. VLLR uses LLMs to decompose tasks into verifiable subtasks and then VLMs to estimate progress to initialize the value function for a brief warm-up phase, avoiding prohibitive inference cost during full training; and self-certainty provides per-step intrinsic guidance throughout PPO finetuning. Ablation studies reveal complementary benefits: VLM-based value initialization primarily improves task completion efficiency, while self-certainty primarily enhances success rates, particularly on out-of-distribution tasks. On the CHORES benchmark covering mobile manipulation and navigation, VLLR achieves up to $56\%$ absolute success rate gains over the pretrained policy, up to $5\%$ gains over state-of-the-art RL finetuning methods on in-distribution tasks, and up to $10\%$ gains on out-of-distribution tasks, all without manual reward engineering. Additional visualizations can be found \href{https://silongyong.github.io/vllr_project_page/}{here}.
\end{abstract}

\section{INTRODUCTION}
\looseness = -1

Long-horizon robotic tasks require a robot to compose multiple skills before receiving meaningful supervision. 
In mobile manipulation~\cite{ehsani2024spoc, hu2024flare}, for example, a robot may need to navigate across rooms, search for an object, reposition obstacles, and finally grasp the target. 
While recent foundation policies trained via large-scale imitation learning~\cite{kim2024openvla, ehsani2024spoc} demonstrate strong individual skills, they often struggle when these skills must be composed over extended horizons. 
Errors accumulate, and sparse end-of-task success signals provide little guidance for correcting intermediate mistakes.

We study how to effectively finetune pretrained foundation policies for long-horizon tasks using reinforcement learning.
Although RL provides a natural mechanism for policy improvement through interaction~\cite{hu2024flare, cui2025grove, rocamonde2023vision, ma2023eureka}, its effectiveness critically depends on reward design~\cite{singh2009rewards}. 
In long-horizon robotics, relying solely on sparse success signals leads to severe credit assignment problems, while manually engineered dense rewards are brittle and task-specific. 
Moreover, reward signals must remain grounded in perception and physical interaction, making scalable reward modeling fundamentally challenging.

We argue that reward design for foundation policy fine-tuning in long-horizon settings must satisfy two requirements:
(1) provide semantic supervision at the level of subgoals to alleviate sparse credit assignment; and 
(2) provide dense per-step feedback to guide local action refinement.

To satisfy the first requirement, we reconceptualize reward modeling in long-horizon robotics as a problem of \textit{exposing and leveraging latent task hierarchy}.
Rather than relying on sparse terminal supervision or manually engineered shaping signals, we explicitly model the semantic structure that underlies compositional tasks.
Long-horizon behavior is inherently hierarchical: successful execution emerges from completing a sequence of intermediate subgoals that progressively approach task goals.


\looseness=-1
Instead of manually defining subgoals, we use a large language model (LLM) to extract intermediate objectives from high-level task descriptions.
Crucially, this decomposition is conditioned on scene-graph representations of the environment, enabling globally consistent and perception-aware planning rather than purely textual reasoning.

\looseness=-1
We then turn the subgoals into explicit intermediate objectives that provide structured supervision during reinforcement learning.
We employ a vision-language model (VLM) to evaluate the alignment between the agent’s observation and each subgoal, yielding a progress estimate that reflects task advancement, which is subsequently turned into a reward that encourages behaviors that complete the subgoals.

To satisfy the second requirement, we introduce \textit{policy self-certainty}~\cite{kang2025scalable}, i.e. a metric that quantifies the confidence of the action distribution, as an intrinsic reward signal.
We conceptualize long-horizon control as an implicit reasoning process embedded within pretrained foundation policies.
Analogous to long-chain reasoning in large language models, sequential decision-making requires maintaining coherent internal representations over extended horizons.
Recent work shows that self-certainty strongly correlates with reasoning correctness in language models~\cite{kang2025scalable}.
Inspired by prior work that justifies the existence of an implicit world model inside a general policy~\cite{richens2025general}, we propose the following insight: \textit{self-certainty serves as a proxy for whether the policy’s latent world model is internally consistent and progressing toward task completion}.
This establishes a principled mechanism for leveraging knowledge already encoded within foundation policies, rather than relying solely on manually engineered external rewards.
To prevent overconfidence in erroneous behaviors, we combine this intrinsic signal with the sparse task-success reward during fine-tuning. In practice, the task-success reward is assigned a larger weight in the overall reward function, ensuring that the optimization objective remains anchored to the true task goal.


Building on these principles, we introduce \underline{V}ision-\underline{L}anguage-\underline{L}ong-horizon \underline{R}eward (VLLR), a structured reward framework for fine-tuning foundation policies in long-horizon robotic tasks.
VLLR combines grounded subgoal decomposition and intrinsic self-certainty to enable scalable and generalizable policy refinement.

We evaluate VLLR on the CHORES benchmark~\cite{ehsani2024spoc}, which features compositional mobile manipulation and navigation tasks.
On in-distribution tasks, VLLR improves absolute success rates by up to $56\%$ over the pretrained foundation policy and consistently outperforms state-of-the-art RL fine-tuning methods based solely on sparse rewards.
Importantly, VLLR generalizes to out-of-distribution task compositions, achieving up to $10\%$ higher success rates than prior methods.
These results demonstrate that structured semantic distillation combined with intrinsic progress modeling provides a practical pathway for extending foundation model fine-tuning to long-horizon robotic control.

In summary, we make the following contributions:
\begin{itemize}
    \item We propose two key principles for reward modeling in long-horizon robotic fine-tuning which guide the design of our reward model: 
    (i) leveraging task hierarchy to provide subgoal-level supervision, and 
    (ii) exploiting intrinsic policy signals to obtain dense per-state feedback.

    \item We propose \textbf{Vision-Language Long-horizon Reward (VLLR)}, which uses VLM-derived semantic progress for \emph{value initialization} and policy self-certainty as an intrinsic reward for stable long-horizon fine-tuning.

    \item On CHORES, VLLR, without manual engineering, achieves up to $56\%$ absolute success rate gains over the pretrained policy and consistently outperforms state-of-the-art RL fine-tuning methods using sparse rewards.
\end{itemize}

\section{Related Work}
\label{sec:related_work}

\subsection{Robotic Foundation Models}
\looseness=-1
Inspired by recent success in developing foundation models in language~\cite{achiam2023gpt} and vision~\cite{liu2023visual}, prior work has developed foundation models that unify language, vision and action in robotics~\cite{kim2024openvla, black2024pi_0, bjorck2025gr00t, ehsani2024spoc, brohan2022rt, zitkovich2023rt, huang2023embodied}. They share a similar training recipe that relies on large scale data and imitation learning, which produces generalizable policy models across diverse tasks. However, they still face challenges when dealing with long-horizon complex scenarios~\cite{fan2025long}. Our work addresses this issue by developing finetuning signal for better generalizability of these policies.

\subsection{Reinforcement Learning for Foundation Models}
Classical reinforcement learning approaches such as policy gradient~\cite{sutton1999policy, schulman2015trust, schulman2017proximal}, value learning~\cite{mnih2013playing,fujimoto2018addressing,haarnoja2018soft}, and preference-based RL~\cite{JMLR:v18:16-634,knox2009interactively,knox2012reinforcement,bradley1952rank} have gained great traction in fine-tuning foundation models, especially LLMs~\cite{bai2022training,rafailov2023direct, guo2025deepseek}. Recently, there has been success in using RL to finetune foundation policies that incorporate vision, language, and actions~\cite{hu2024flare, lu2025vla}. Unlike prior approaches that rely on sparse rewards or costly reward model training from labeled data, our method leverages foundation models in a zero-shot manner to provide subgoal decomposition and generalizable extrinsic reward signals for robotic fine-tuning.

\subsection{Reward Design for Robotic Tasks}
Traditional reinforcement learning (RL) relies heavily on carefully specified reward functions to guide agent behavior. In robotics, this is particularly challenging because of the sparse task success signals and high-dimensional continuous control. Poorly designed rewards often lead to misaligned incentives, suboptimal behaviors, or even unsafe strategies.

\looseness=-1
Recently, the rise of foundation models (FMs) has opened new directions for addressing the Reward Design Problem~x\cite{singh2009rewards}. Several works have explored how pretrained models can be leveraged to generate or shape reward functions~\cite{ma2023eureka, rocamonde2023vision, cui2025grove, singh2009rewards, ma2024vision}. For example, Eureka~\cite{ma2023eureka} uses LLMs to define reward functions by writing code, whereas VLM-RM~\cite{rocamonde2023vision} demonstrates that VLM-based similarity score can serve as a reward signal for repetitive agent patterns such as jump and walk. Leveraging the capabilities and generalizability of foundation models, this paradigm reduces reliance on hand-crafted shaping reward and mitigates the limitations of purely sparse reward signals.

Despite these advances, FM-based reward design still remains an open challenge for long-horizon tasks~\cite{lin2024navigating}. Prior works~\cite{hu2024flare, lu2025vla} either rely on sparse reward signals or a trained reward model, both of which have limited scalability and generalizability since they either cause sample inefficiency or require task-specific manual-engineering. To address these limitations, we have developed a generalizable reward model that uses foundation models inspired by their usage in short-horizon or repetitive skills~\cite{ma2023eureka, cui2025grove, rocamonde2023vision}.

\section{Method}

\subsection{Method Overview}
\looseness=-1
Our goal is to fine-tune a pretrained robotic foundation policy using reinforcement learning algorithms such as PPO for long-horizon mobile manipulation and navigation tasks. 
The pretrained policy maps observations to a stochastic action distribution over the robot's action space.
Since PPO requires both a policy and a value function, and the pretrained policy obtained from imitation learning does not provide a suitable critic, we introduce a value function following FLaRe~\cite{hu2024flare} to enable PPO-based fine-tuning.

\looseness=-1
The overall pipeline of VLLR consists of three key components and a two-stage optimization procedure.

\looseness=-1
First, we construct a progress signal at the subgoal level, meaning that the progress increases whenever a subgoal is achieved and eventually reaches its maximum when the overall task is completed. We call it \textit{extrinsic reward} since it'll be provided by external foundation models like LLMs and VLMs.
Given a task instruction and a structured scene graph of the whole environment, a large language model (LLM) decomposes the task into an ordered sequence of grounded subgoals.
A vision-language model (VLM) then evaluates the agent’s visual observations with respect to the ordered subgoals, determining which subgoal the agent is currently attempting and estimating the resulting progress toward the overall task.
This provides coarse-grained supervision that reflects task-level advancement.

\looseness=-1
Second, we introduce an intrinsic reward signal derived from the pretrained policy itself.
We measure policy self-certainty, which quantifies the concentration of the action distribution. 
Following recent work suggesting that large pretrained policies implicitly encode a form of world model, i.e., internal representations capturing regularities of environment dynamics and task structure, we interpret action distribution concentration as a proxy for how well the current state aligns with this internal model of task progression. 
The intrinsic reward is dense and provided in a per-step manner, which is complementary to the extrinsic reward which is coarse-grained.

\looseness=-1
Third, to ensure computational efficiency, we adopt a two-stage training procedure.
In Stage I, the extrinsic reward, i.e., VLM-derived progress signal, is used to initialize the value function.
In Stage II, the policy is fine-tuned using PPO with the intrinsic reward, i.e., self-certainty, and sparse task success rewards, without further VLM queries.

\looseness=-1
Together, these components form a hierarchical reward design:
subgoal supervision provides coarse-level structure, intrinsic self-certainty provides fine-grained action-level guidance and sparse task success reward anchors the optimization towards the true task goal.
An overview of the full pipeline is shown in Fig.~\ref{fig:full}.

\subsection{Extrinsic Reward Design}
\looseness=-1
In order to construct the reward signal that encourages progress made in completing subgoals and towards the overall goal, we adopt a four-step design. We first use LLMs to do task decomposition to provide subgoals. Secondly, the subgoals are sent to VLMs to determine which subgoal the agent is doing and estimate the overall progress. Third, we correct the raw VLM progress estimation due to its noisy nature. Fourth, we use the corrected progress estimation only at the first stage of training, i.e. value function initialization.

\subsubsection{Task Decomposition with Large Language Models}
\label{subsec:subtask}
\looseness=-1
Long-horizon robotic tasks like mobile manipulation are typically specified by a natural language instruction $\mathcal{I}$, which is often ambiguous and lacks step-by-step guidance for execution. In reinforcement learning, this ambiguity translates into a sparse-reward setting, where only terminal success signals are available, making credit assignment difficult.

To mitigate this issue, we introduce a structured task decomposition module based on a large language model (LLM). Formally, the LLM operates as a mapping
\[
\text{LLM}: (\mathcal{G}, \mathcal{I}) \rightarrow \{g_1, g_2, \dots, g_K\},
\]
\looseness=-1
where $\mathcal{G}$ is the structured scene graph describing global environment information (e.g., rooms and objects for navigation tasks, and object relations for manipulation tasks), $\mathcal{I}$ is the high-level task instruction, and $\{g_k\}_{k=1}^K$ is an ordered sequence of subgoals constituting a task decomposition.

The key advantage is that the LLM has access to global structural information through $\mathcal{G}$, allowing it to reason over the entire environment rather than relying on partial observations. As a result, it can generate a coherent plan that progressively reduces spatial and semantic uncertainty.

\begin{figure}[h]
\begin{center}
\includegraphics[width=0.95\linewidth]{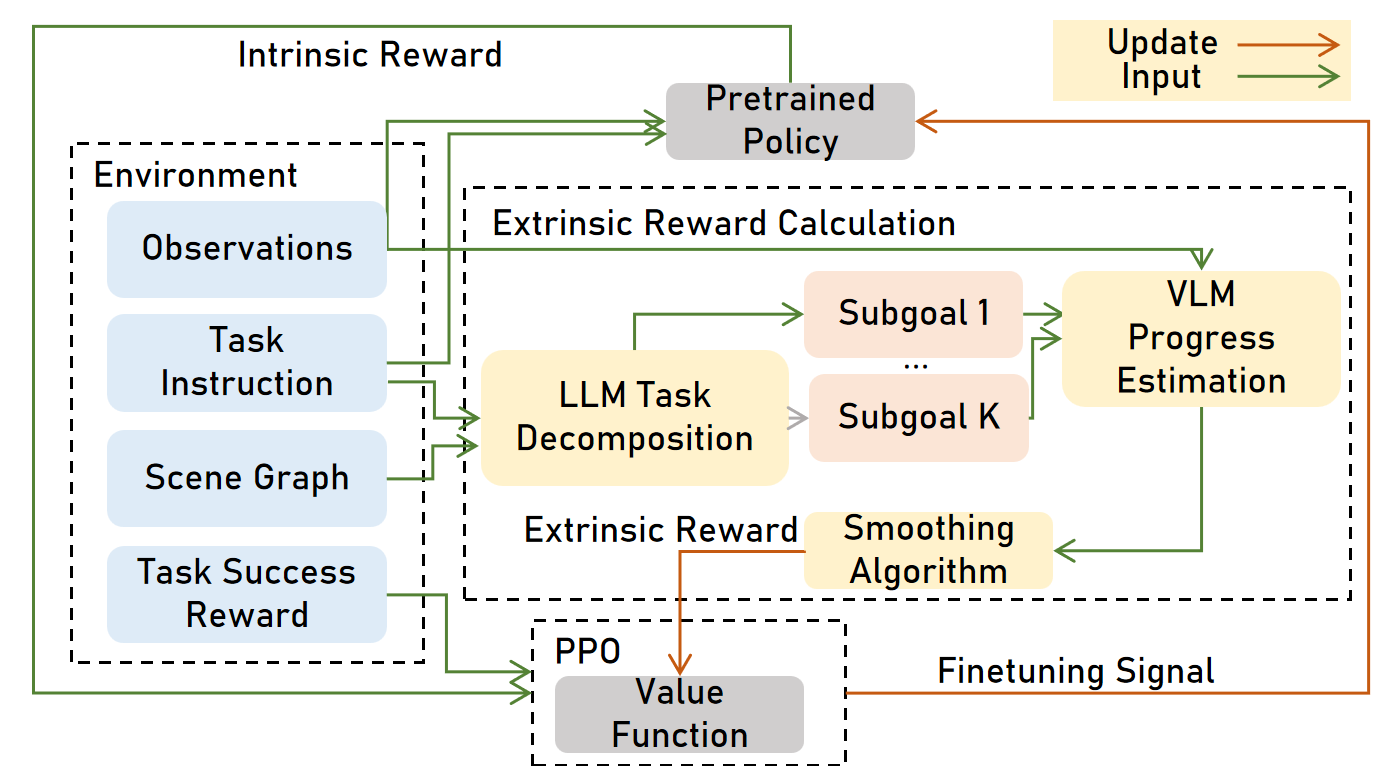}
\end{center}
\caption{Method overview. VLLR is a reward model with three components, i.e. sparse task success reward, intrinsic reward and extrinsic reward (Sec.~\ref{subsec:final}). The environment provides task instruction and a scene graph context to the LLM for task decomposition (Sec.~\ref{subsec:subtask}). Then the decomposed subgoals are fed into a VLM together with the current and last observations (Sec.~\ref{subsec:vlm_pe}). The VLM provides a noisy progress estimate which is then smoothed out (Sec.~\ref{subsec:smooth}) and used as a reward signal for initializing the value function (Sec.~\ref{subsec:value}) used for PPO. The intrinsic reward is calculated using the policy's output action distribution (Sec.~\ref{subsec:intrinsic}). It is fed together with the task success signal into the PPO algorithm and then produces updates for finetuning the pretrained policy (Sec.~\ref{subsec:final}).}
\label{fig:full}
\end{figure}

As shown in Fig.~\ref{fig:full}, each subgoal / subtask localizes the objective and provides an intermediate verification point, effectively transforming a long-horizon sparse-reward problem into a sequence of structured short-horizon objectives.

We implement this module using a fixed prompt template that incorporates both the scene graph and the task instruction. We'll release the full prompt upon acceptance.

\subsubsection{Progress Estimation with Vision-Language Models}
\label{subsec:vlm_pe}

\looseness=-1
Given the task decomposition $\{g_k\}_{k=1}^K$ generated by the LLM, we next transform the plan of subgoals into a learning signal that encourages subgoal completion.
The key intuition is: if the agent's observation indicates that a subgoal has been achieved or that the agent is closer to the next subgoal, then the agent has made progress toward the overall task.

To operationalize this idea, we employ a vision-language model (VLM) to verify subgoal completion from visual observations.
Given the ordered subgoals and the current observation, the VLM explicitly determines which subgoal the agent is currently attempting and how much progress has been made toward completing it.
The VLM's output is then transformed into a scalar score that reflects the agent's progress toward the overall task, which increases as the agent successfully completes successive subgoals.

\looseness=-1
We formulate the usage of the VLM as a mapping
\[
\text{VLM}: (\{g_k\}_{k=1}^K, o_{t-1}, o_t) \rightarrow p_t,
\]
\looseness=-1
where $\{g_k\}_{k=1}^K$ is the ordered subgoal sequence, $o_{t-1}, o_t$ represents the two consecutive observations at time $t-1$ and $t$, and $p_t \in [0,1]$ is a scalar progress estimate indicating how much the agent has advanced toward the overall task.

\begin{figure}[h]
\begin{center}
\includegraphics[width=0.95\linewidth]{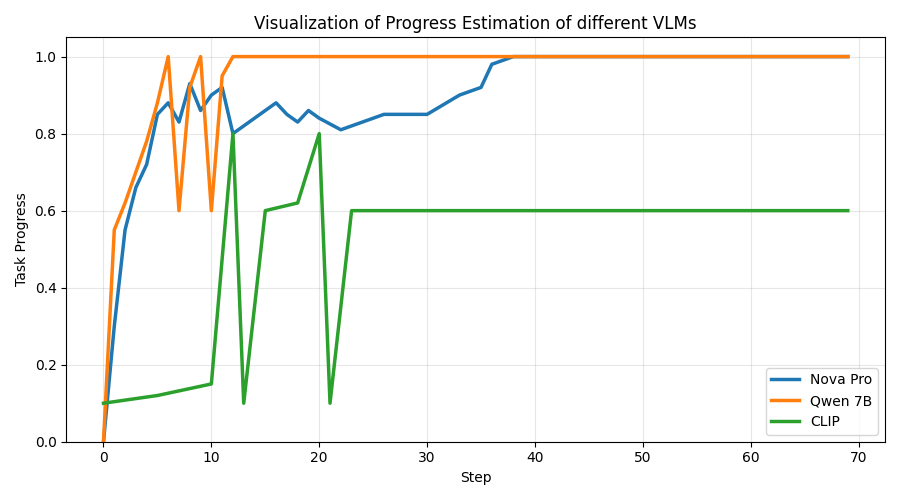}
\end{center}
\vspace{-15pt}
\caption{We showcase an example to compare different VLM progress estimation including Qwen, Nova Pro and CLIP. The VLMs are tasked to estimate progress for finding a laptop that is not in plain sight. We found that Qwen tends to over-saturate the estimation, mostly caused by falsely recognizing the laptop, CLIP is unstable, wrong and unable to identify overall task completion, and Nova Pro is the best at progress estimation, which means it provides correct signal when seeing the laptop and approaching it. Here the rollout is collected using A* and the ground truth progress should be linearly increasing.}
\label{fig:vlm_comparison_heuristics}
\end{figure}

\looseness=-1
Through experimentation, we found that Nova Pro, a VLM, tends to provide a progress signal that follows a similar progression trend as the ground-truth task completion rate, although it may fluctuate across timesteps due to perception noise. We made this observation by comparing several VLMs. Specifically, we observe that CLIP tends to show no correlation with the task progress, Qwen tends to predict task completion early, whereas Nova Pro tends to predict task completion at a later time and its prediction is gradually increasing when looking at an offline video recorded using A* as provided in the CHORES benchmark~\cite{ehsani2024spoc}.
We provide a representative example in Fig.~\ref{fig:vlm_comparison_heuristics}. We adopt Nova Pro as the VLM for our method.

\looseness=-1
To derive the reward for VLLR from the progress estimation from Nova Pro, we introduce a \emph{running maximum progress} variable $\hat{p}_t$, which tracks the highest progress estimate observed so far:
\[
\hat{p}_{t+1} = \max(\hat{p}_t, p_{t+1}).
\]

\looseness=-1
We then define the reward as the incremental improvement in this running maximum:
\[
R_{VLM}(s_t, a_t, s_{t+1}) = \hat{p}_{t+1} - \hat{p}_t.
\]

This formulation ensures that rewards are only assigned when the agent achieves new progress.
However, it also makes the reward sensitive to occasional spurious high progress estimates (e.g., hallucinated peaks): once an erroneously large $p_t$ is observed, the running maximum $\hat{p}_t$ saturates and suppresses future rewards.
We address this \emph{premature saturation} issue in the third step by stabilizing the progress estimates before computing the running maximum.

\begin{algorithm}
\caption{Premature saturation prevention for correcting VLM progress estimation}
\label{algo:post-process}
\begin{algorithmic}[1]   
\STATE Initialize size of window $S$, pruning threshold $T$, buffer $B$ that contains all progress estimations $\{p_{1:n}\}$, a new buffer that contains all processed rewards $\{r_{1:n}\}$, a running maximum progress $RMP = 0$
\FOR{$i = 1$ to $n$}
    \STATE $W = \{p_{i-S:i+S}\} \setminus \{ p_i \}$
    \STATE $M = \operatorname{Median}(W)$
\IF{$p_i-M > T$}
    \STATE $r_i =0$ 
\ELSE
    \STATE $r_i = p_i$
\ENDIF

\ENDFOR
\FOR{$i = 1$ to $n$}
\IF{$r_i > RMP$}
    \STATE $temp = r_i$
    \STATE $r_i =r_i - RMP$
    \STATE $RMP = temp$
\ELSE
    \STATE $r_i = 0$
\ENDIF

\ENDFOR
\RETURN $\{r_{1:n}\}$
\end{algorithmic}
\end{algorithm}

\subsubsection{Preventing Premature Saturation of Running Maximum Progress}
\label{subsec:smooth}
\looseness=-1
Once an erroneously large progress value is produced, normally due to false object recognition caused by hallucinations, it increases the \emph{running maximum progress} $\hat{p}_t$, suppressing future rewards and effectively removing learning signals for subsequent steps.

To address this issue, we stabilize the progress estimates from Nova Pro at the trajectory level before the PPO update. Specifically, after collecting a rollout, we retrieve the sequence of progress predictions
\[
\{p_1, p_2, \dots, p_T\},
\]
from the PPO buffer and apply temporal consistency filtering using a sliding window of size $w$ (e.g., $w = 5$). Genuine task progress typically evolves smoothly over time, whereas hallucinated predictions appear as isolated spikes.

For example, consider the sequence
\[
\{0.1,\, 0.1,\, 0.9,\, 0.2,\, 0.2\}.
\]
The value $0.9$ is inconsistent with both its preceding and succeeding neighbors and is therefore identified as a spurious peak and corrected using median filtering.

\begin{figure}[h]
\begin{center}
\includegraphics[width=0.95\linewidth]{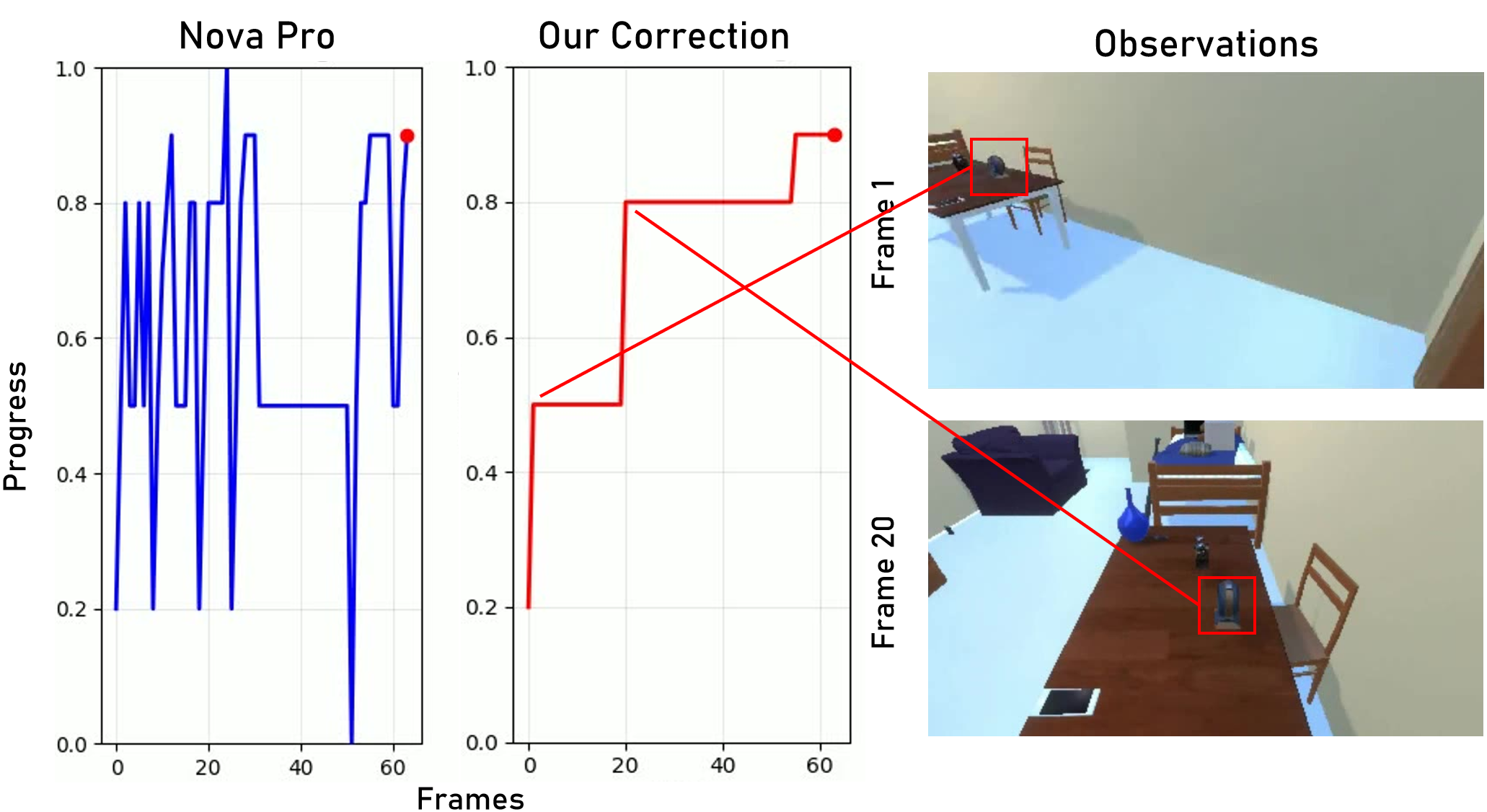}
\end{center}
\caption{We showcase an example where raw progress estimation from Nova Pro is noisy and our method is able to identify the actual progress made by the agent. The task is to find a clock and grasp it. In the first column, we can see that throughout the rollout, the progress estimation is noisy and problematic, unable to provide meaningful reward signals. In the second column, our method is able to provide three clear jumps in terms of progress estimation. The first jump rewards the fact that the robot sees the clock, the second jump rewards the robot positioning itself in front of the clock and the third jump rewards the robot actually picking up the clock. In the third column, we showcase the actual observation Nova Pro is provided. We highlight the clock in red box in frame 1 and 20. Frame 1 showcase the fact that the robot sees the clock and frame 20 demonstrates the robot positioning itself in front of the clock. We connected these two observations to the first two jumps in our progress estimation.}
\label{fig:vlm_comparison}
\vspace{-15pt}
\end{figure}

\looseness=-1
Since this correction is applied after rollout collection and before the PPO update, we can leverage both past and future steps within the stored trajectory to reliably identify and suppress such artifacts. The complete procedure is detailed in Algo.~\ref{algo:post-process}. We also showcase an example in Fig.~\ref{fig:vlm_comparison} where the raw progress estimation from Nova Pro is noisy, whereas the estimation after our correction is able to provide clear reward signals when visually verifiable cues are provided, e.g. in the figure the clock is correctly identified and Nova Pro recognizes the fact that the agent is moving closer towards the clock. We can then use this smoothed progress estimation for computing the reward as discussed in Sec.~\ref{subsec:vlm_pe}. 

\subsubsection{VLM Reward as Proxy for Value Function Initialization}
\label{subsec:value}

We observe that directly querying a VLM, e.g. LLM-based VLM like Nova Pro, at every environment step is computationally prohibitive, especially in long-horizon settings due to high latency and inference cost.

We instead use the VLM only during a short initialization phase to supervise the value function. After imitation pretraining, the VLM-derived progress signal is used to initialize the value function for 200K steps ($<0.5\%$ of 50M total steps). No further VLM queries are performed afterward.

Unlike prior work that optimizes policies against VLM rewards throughout training~\cite{rocamonde2023vision, cui2025grove}, we only use the VLM reward at Stage~I for value function initialization, which then provides dense value estimates at negligible cost.


\subsection{Intrinsic Reward Design}
\label{subsec:intrinsic}
\looseness=-1
Beyond leveraging extrinsic subgoal-level supervision, we further exploit intrinsic signals from the policy model itself for per-step supervision. Recent advances in reinforcement learning for large language models (LLMs) have shown that intrinsic confidence-related metrics can serve as effective optimization signals for improving reasoning performance~\cite{kang2025scalable, zhao2025learning}. In robotic policy learning, prior work~\cite{zhang2024grape} primarily relied on log-probability-based intrinsic objectives, where higher likelihood assigned to executed actions was directly used as a reward signal.

Following these insights, we adopt \emph{self-certainty}~\cite{kang2025scalable} as an intrinsic reward signal for robotic policy finetuning. Rather than directly maximizing the log-probability of selected actions, self-certainty measures the overall concentration of the policy distribution. Specifically, it reflects how peaked the action distribution is: when the policy assigns high probability to a small set of actions, the distribution has lower entropy and higher self-certainty, indicating stronger confidence in the selected behavior.

Formally, self-certainty is defined as:
\begin{equation}
    R_{SC} = -\frac{1}{|A|}\sum_{i=1}^{|A|}\log (|A|\cdot \pi_\theta(i)),
    \label{eq:scr}
\end{equation}
where $|A|$ denotes the size of the discrete action space and $\pi_\theta(i)$ is the probability assigned by the policy to action $i$.


\looseness=-1
While prior work established a positive correlation between self-certainty and reasoning correctness in LLMs~\cite{kang2025scalable}, we extend this perspective to robotic foundation models from a structural viewpoint. This makes intuitive sense since recent studies~\cite{richens2025general} suggest that large pretrained policies implicitly encode a form of internal world model, capturing regularities of environment dynamics and task structure within their representations. 

\looseness=-1
Under this view, self-certainty is not merely a measure of the concentration of the output distribution, but reflects the internal belief of how the world would evolve based on the policy’s latent world representation. When the observation aligns well with the policy’s internal model of task progression, the action distribution tends to be more concentrated.



\subsection{Final Reward Model}
\label{subsec:final}
\looseness=-1
Combining the above components, we obtain the final reward model under our two-stage optimization framework:

\begin{equation}
\begin{aligned}
    R_{\text{full}} &=
    \alpha \, \mathbb{1}_{\{\text{Stage I}\}} \, R_{VLM}
    + \beta \, \mathbb{1}_{\{\text{Stage II}\}} \, R_{SC}
    \\ &+ \phi \, \mathbb{1}_{\{\text{Stage II}\}} \, R_{task},
\end{aligned}
    \label{eq:full_r}
\end{equation}

where $\alpha, \beta, \phi$ are weighting coefficients for different components. 
Here, $\mathbb{1}_{\{\text{Stage I}\}}$ and $\mathbb{1}_{\{\text{Stage II}\}}$ are indicator functions that activate the corresponding reward terms only during the specified training stage. Specifically, $\mathbb{1}_{\{\text{Stage I}\}} = 1$ during the \emph{value initialization stage} and $0$ otherwise, while $\mathbb{1}_{\{\text{Stage II}\}} = 1$ during the \emph{policy finetuning stage} and $0$ otherwise.

\looseness=-1
In Stage~I, the VLM-derived reward $R_{VLM}$ is used solely to initialize the value function, which serves as a value prior that incorporates knowledge of task hierarchy. In Stage~II, the optimization relies only on the intrinsic self-certainty reward $R_{SC}$ and the sparse task completion signal $R_{task}$.

\begin{table*}[t]
\centering
\footnotesize
\caption{Success and Episode-length weighted Success (SEL) on the CHORES~\cite{ehsani2024spoc} benchmark. Our method outperforms the previous SoTA methods by providing generalizable reward signal across tasks. Baselines are chosen from FLaRe~\cite{hu2024flare}.}
\begin{tabular}{l||c|c|c|c|c|}  
\toprule
\multirow{2}{*}{Success (SEL)} 
    & \multicolumn{1}{c|}{\textbf{IL+RL: Dense Reward}}
    & \multicolumn{3}{c|}{\textbf{IL+RL: Sparse Reward}} 
    & \multicolumn{1}{c|}{\textbf{IL Only}} \\ 
\cmidrule(lr){2-2} \cmidrule(lr){3-5} \cmidrule(lr){6-6}
    & VLLR (Ours) & FLaRe & PIRLNav & JSRL & SPOC \\ 
\midrule

ObjectNav 
& \textbf{87.6 (77.7)}
    & 85.0 (67.6)
    & 20.0 (7.0)  
    & 21.0 (15.6) 
    & 55.0 (42.2) \\

Fetch  
& \textbf{70.7 (63.2)}
    & 65.23 (54.7)
    & 0.0 (0.0)
    & 2.9 (2.8) 
    & 14.0 (10.5)\\

PickUp 
& \textbf{97.0 (96.4)}
    & 91.8 (90.4)
    & 0.0 (0.0) 
    & 50.9 (47.7) 
    & 90.1 (86.9) \\

RoomVisit 
& \textbf{68.3} (64.8)
    & 60.87 (\textbf{65.7}) 
    & 12.5 (11.0)  
    &  19.0 (18.6) 
    & 40.5 (35.7) \\

\bottomrule
\end{tabular}
\label{tab:indist}
\vspace{-1em}
\end{table*}

\begin{table}[t]
\centering
\scriptsize
\setlength{\tabcolsep}{0.5pt}
\caption{Our method can serve as the reward signal for out-of-distribution tasks that are never seen by the base model, and make the policy achieve state-of-the-art performance. Baselines are chosen from FLaRe~\cite{hu2024flare}}
\begin{tabular}{c||c||cccc} 
\toprule
                        Success (SEL)& VLLR (Ours) & FLaRe    & Poliformer (Sp)                            & SPOC++                    & Poliformer (De)                         \\
 \midrule
 
 ObjNavRel       
 & \textbf{67.4 (61.5)} & 66.3 (60.6)  &  6.7 (6.7)  & 54.5 (44.6)
 &  36.1 (32.4)   
 \\ 
 
 ObjNavAff      
 
 & \textbf{90.1 (78.2)} & 79.7 (70.6)  &    35.5 (29.4)     & 62.4 (50.6)
 &    53.8 (43.1)      
 \\ 

 \bottomrule
\end{tabular}
\label{tab:ood}
\end{table}

\begin{table}[t]
\centering
\scriptsize
\setlength{\tabcolsep}{4pt}
\caption{Ablation study on three representative tasks. We exclude PickUp (near-ceiling at 97\%) and RoomVisit (SEL characteristics discussed in Sec.~\ref{sec:indist}) as they offer limited signal for component-level comparison. Base: sparse reward only; SCR: self-certainty reward; VLM: VLM progress reward; Full: complete VLLR.}
\begin{tabular}{c||cccc} 
\toprule
                        Success (SEL)& Base & SCR    & VLM                            & Full                                           \\
 \midrule
 
 Fetch       
 & 65.23 (54.7) & 69.18 (60.4) & 68.00 (63.8)  &  70.7 (63.2)
  
 \\ 
 ObjectNav      
 
 & 85.0 (67.6) & 86.7 (76.6) & 87.1 (77.5)    &  87.6 (77.7)
  
 \\ 
  ObjNavAff      
 
 & 79.7 (70.6) & 89.7 (76.5) &  89.0 (77.1) &  90.1 (78.2)
     
 \\ 
 
     

 \bottomrule
\end{tabular}
\label{tab:abl}
\vspace{-1em}
\end{table}





\section{Experiments}
We conduct our experiments on the CHORES benchmark~\cite{ehsani2024spoc}, which consists of long-horizon robotic tasks that mainly focus on mobile manipulation. 
We observe that our method effectively guides policy optimization, resulting in both higher success rates and improved efficiency compared to baselines. Our method also enables generalization to unseen tasks in the pretraining dataset.

\subsection{Implementation Details}
\looseness=-1
We follow the same settings in FLaRe~\cite{hu2024flare} for the PPO hyperparameters. For the weighting factors in Eq.~\ref{eq:full_r}, we set $\alpha = 1$, $\beta = 0.1$ and $\phi = 10$. Value function learning is done for 200K steps. For Object Navigation and RoomVisit we run 20M steps for training, and for the rest of the two in-distribution tasks, i.e. Fetch and Pick-up, we run 50M steps since it requires fine-grained manipulation. Object Relative Attributes Navigation and Object Affordance Navigation are trained for 50M steps.
The threshold for noise detection is set as the reciprocal of the number of subtasks for the specific task and the window size is set as 9. For task decomposition we use Claude-3.7-Sonnet. For progress estimation we use Amazon Nova Pro, a lighter-weight but also best-performing VLM selected to minimize per-step inference latency: since the VLM is queried at every environment step during Stage~I rollout collection, inference speed directly impacts training throughput.

\subsection{Benchmark Configuration}
\looseness=-1
The CHORES benchmark defines an observation space consisting of two egocentric RGB cameras (resolution $384 \times 224$) facing orthogonal directions, i.e., one aligned with the navigation direction and the other with the manipulator arm. At the beginning of each episode, a natural language instruction is sampled and appended to the observation, specifying the intended task. The policy has an action space of 20 discrete actions, including base translation, base rotation, arm translation, gripper rotation, as well as pickup, dropoff, task completion signal, and episode termination. Task specifications covering a diverse set of household activities. Episodes terminate either when a task is completed or when the maximum step limit is reached, in which case the attempt is marked as failed. We evaluate performance using two metrics: task \textbf{Success Rate} ($S \in \{0,1\}$ per episode) and \textbf{Success weighted by Episode Length (SEL)}~\cite{eftekhar2023selective}, defined as
\begin{equation}
    \text{SEL} = S \cdot \frac{t_{\min}}{\max(t_{\min},\, t)},
\end{equation}
where $t_{\min}$ is the minimum number of steps required to complete the task, which is calculated using A* and provided by the dataset, and $t$ is the agent's actual episode length. 
The benchmark is built on 191,568 houses from ProcThor~\cite{deitke2022️}, split into training and testing environments at a 10:1 ratio to ensure evaluation on previously unseen houses. 
Please refer to the original paper for details due to space limit~\cite{ehsani2024spoc, hu2024flare}.

\subsection{In-distribution tasks}
\label{sec:indist}
\looseness=-1
We first conduct experiments on four tasks that are in the pretraining dataset of our foundation policy, SPOC~\cite{ehsani2024spoc}, namely Fetch, Pick-up, RoomVisit and Object Navigation. The full task settings can be found in the Sec.~\ref{supp:task_def} in the appendix. We compare our method with the SOTA method FLaRe~\cite{hu2024flare}, the baseline method SPOC~\cite{ehsani2024spoc}, and PIRLNav~\cite{ramrakhya2023pirlnav} and JSRL~\cite{uchendu2023jump}, which apply RL finetuning with sparse task-success rewards. As shown in Tab.~\ref{tab:indist}, our method achieves up to $5\%$ absolute success rate improvement over the state-of-the-art (SOTA) method. Importantly, this gain is obtained without relying on task-specific reward engineering or privileged environment annotations. We also notice significant improvements in terms of the policy's efficiency in completing the task. Compared to the baseline, the policy trained with our reward model is able to provide up to 10\% absolute efficiency improvement, which demonstrates that our method is able to provide informative information about the task progress and the agent's behavior. We observe that on RoomVisit, FLaRe achieves a slightly higher SEL. We hypothesize that RoomVisit primarily emphasizes visiting multiple spatial targets, where global route ordering plays a larger role than fine-grained action efficiency. In such scenarios, small variations in path ordering can slightly affect SEL, even when overall behavioral efficiency remains comparable.

\subsection{Out-of-distribution tasks}
We further study whether the reward model we designed is generalizable enough to guide the adaptation to unseen task settings of the pretrained model. Specifically, we consider two tasks that are not in the pretraining data, namely Object Affordance Navigation and Object Relative Attribute Navigation. The full details can also be found in Sec.~\ref{supp:task_def} in the appendix. For OOD tasks we additionally compare against Poliformer~\cite{ehsani2024spoc} and SPOC++~\cite{ehsani2024spoc}, which leverage privileged information in the form of human-defined task-specific dense rewards and were evaluated on ObjNavRel and ObjNavAff in the SPOC benchmark; PIRLNav and JSRL are excluded here because FLaRe has  superior performance. As shown in Tab.~\ref{tab:ood}, the experimental results demonstrate that our method is able to provide superior information regardless of the task setting and is able to guide the policy model to better adapt to the task goal and provides up to $10\%$ absolute performance gain compared to SOTA on the OOD tasks.

\subsection{Analysis}
\label{sec:analysis}
To better understand the roles of different reward components, we perform detailed ablation studies.

\paragraph{Effect of Self-Certainty on Optimization Dynamics.}

We observe that incorporating self-certainty significantly accelerates convergence: the best-performing checkpoint appears much earlier than with sparse reward alone (e.g., ~30M vs. 50M steps in our setting).

This suggests that self-certainty provides a dense intrinsic signal that improves credit assignment during finetuning. Given that the pretrained policy already encodes substantial task structure from imitation learning, self-certainty reinforces internally consistent predictions and speeds up alignment toward successful behaviors.

However, assigning an overly large weight to $R_{SC}$ destabilizes training. Excess confidence can override existing value estimates, leading to catastrophic forgetting and performance collapse. Hence, careful balancing between intrinsic certainty and task-level supervision is essential.

\paragraph{Effect of VLM-Based Value Initialization.}

While self-certainty primarily improves convergence speed and final success rate, VLM-based value initialization contributes more significantly to task completion efficiency. Policies initialized with VLM supervision tend to require fewer environment steps to complete successful episodes.

\looseness=-1
We attribute this to the semantic grounding provided during Stage I. The VLM-derived progress signal encodes high-level task structure, encouraging trajectories that follow the plans. As a result, the learned value function better reflects global task progression, reducing unnecessary exploration and redundant movements.

\paragraph{Success vs. Failure Behavior Patterns.}

\looseness=-1
Qualitatively, policies trained with sparse reward alone tend to exhibit indecisive exploration and delayed commitment to task-relevant regions. Self-certainty-enhanced policies qualitatively appear to produce more decisive action sequences, and when combined with VLM initialization, trajectories qualitatively align more closely with task decomposition structure.
These observations suggest complementary roles for the two components: self-certainty appears to enhance internal consistency and optimization speed, while VLM initialization appears to improve global task efficiency through semantic grounding. Quantitative trajectory analysis is left to future work.

\paragraph{Limitations.}
Our evaluation is conducted on the CHORES benchmark~\cite{ehsani2024spoc}, which is also the sole benchmark used in FLaRe~\cite{hu2024flare}, the primary state-of-the-art comparison and a recent ICRA 2025 accepted paper. CHORES spans a diverse range of long-horizon household tasks—from pure navigation (ObjNav, RoomVisit) to dexterous manipulation (PickUp) and combined navigation-and-manipulation (Fetch), together with two out-of-distribution tasks requiring affordance and relational reasoning (ObjNavAff, ObjNavRel). Evaluation is conducted on held-out test houses drawn from a 10:1 split of 191,568 procedurally generated ProcThor houses~\cite{hu2024flare}, yielding approximately 17,000 unseen test environments. The individual components of VLLR: LLM-based task decomposition, VLM progress estimation, and policy self-certainty, rely only on language-conditioned task specifications and egocentric visual observations, and are not specific to this benchmark or action space. Extending VLLR to continuous-action manipulation benchmarks is an important direction for future work.

\section{CONCLUSIONS}
\looseness=-1
In this paper, we present VLLR, a generalizable reward model for finetuning robotic foundation models with RL on long-horizon tasks. VLLR combines two components: (1) an extrinsic reward that measures semantic task progress using the planning and visual verification capabilities of LLMs/VLMs, which is queried only during value learning to reduce compute, and (2) an intrinsic reward derived from the policy’s self-certainty, which correlates with task success. Together, they provide dense, informative signals aligned with sparse success rewards, improving sample efficiency. Experiments on the CHORES dataset demonstrate VLLR’s strong generalization across both in- and out-of-distribution tasks, suggesting a promising direction for unlocking the potential of robotic foundation models.

\addtolength{\textheight}{-0.5cm}   



\vspace{-5pt}
\section*{APPENDIX}

\subsection{Task definition in CHORES}
We refer the reader to the CHORES~\cite{ehsani2024spoc} benchmark for comprehensive understanding of the tasks setting.
\label{supp:task_def}
\looseness=-1
\paragraph{Fetch} 
In the Fetch task, the agent is instructed to both locate and acquire a specified object. For example, \emph{“locate a mug and pick up that mug”}.

\looseness=-1
\paragraph{Pick-Up} 
The Pick-Up task evaluates an agent’s ability to manipulate an object that is already within its line of sight. An example instruction is \emph{“pick up a mug”}. 

\paragraph{ObjNav} 
Object Navigation requires the agent to go to an instance of a specified object category in the environment, such as \emph{“find a mug”}. 

\paragraph{RoomVisit} 
\looseness=-1
RoomVisit focuses on comprehensive environment coverage. For example, \emph{“visit every room in this 5-room house”}. 

\paragraph{ObjNavRel} 
Object Navigation with Relative Attributes extends ObjNav by requiring reasoning over object properties. For instance, \emph{“find the largest apple”} 

\paragraph{ObjNavAff} 
Object Navigation with Affordances requires the agent to interpret instructions that specify functional properties rather than object categories. For example, \emph{“find something I can sit on”}.

\section*{ACKNOWLEDGMENT}
This work was developed during the internships of SY and CQ at Amazon Robotics. SY, YX, KS were funded in part by the Army Research Laboratory (ARL) award W911QX-24-F-0049, DARPA award FA8750-23-2-1015, ONR award N00014-23-1-2840, and ONR MURI grant N00014-25-1-2116.


\bibliographystyle{IEEEtran}
\bibliography{references}

@inproceedings{ehsani2024spoc,
  title={Spoc: Imitating shortest paths in simulation enables effective navigation and manipulation in the real world},
  author={Ehsani, Kiana and Gupta, Tanmay and Hendrix, Rose and Salvador, Jordi and Weihs, Luca and Zeng, Kuo-Hao and Singh, Kunal Pratap and Kim, Yejin and Han, Winson and Herrasti, Alvaro and others},
  booktitle={Proceedings of the IEEE/CVF Conference on Computer Vision and Pattern Recognition},
  pages={16238--16250},
  year={2024}
}

@article{achiam2023gpt,
  title={Gpt-4 technical report},
  author={Achiam, Josh and Adler, Steven and Agarwal, Sandhini and Ahmad, Lama and Akkaya, Ilge and Aleman, Florencia Leoni and Almeida, Diogo and Altenschmidt, Janko and Altman, Sam and Anadkat, Shyamal and others},
  journal={arXiv preprint arXiv:2303.08774},
  year={2023}
}

@article{liu2023visual,
  title={Visual instruction tuning},
  author={Liu, Haotian and Li, Chunyuan and Wu, Qingyang and Lee, Yong Jae},
  journal={Advances in neural information processing systems},
  volume={36},
  pages={34892--34916},
  year={2023}
}

@article{kim2024openvla,
  title={Openvla: An open-source vision-language-action model},
  author={Kim, Moo Jin and Pertsch, Karl and Karamcheti, Siddharth and Xiao, Ted and Balakrishna, Ashwin and Nair, Suraj and Rafailov, Rafael and Foster, Ethan and Lam, Grace and Sanketi, Pannag and others},
  journal={arXiv preprint arXiv:2406.09246},
  year={2024}
}

@article{bjorck2025gr00t,
  title={Gr00t n1: An open foundation model for generalist humanoid robots},
  author={Bjorck, Johan and Casta{\~n}eda, Fernando and Cherniadev, Nikita and Da, Xingye and Ding, Runyu and Fan, Linxi and Fang, Yu and Fox, Dieter and Hu, Fengyuan and Huang, Spencer and others},
  journal={arXiv preprint arXiv:2503.14734},
  year={2025}
}

@article{fan2025long,
  title={Long-VLA: Unleashing Long-Horizon Capability of Vision Language Action Model for Robot Manipulation},
  author={Fan, Yiguo and Ding, Pengxiang and Bai, Shuanghao and Tong, Xinyang and Zhu, Yuyang and Lu, Hongchao and Dai, Fengqi and Zhao, Wei and Liu, Yang and Huang, Siteng and others},
  journal={arXiv preprint arXiv:2508.19958},
  year={2025}
}

@article{black2024pi_0,
  title={$\pi_0 $: A Vision-Language-Action Flow Model for General Robot Control},
  author={Black, Kevin and Brown, Noah and Driess, Danny and Esmail, Adnan and Equi, Michael and Finn, Chelsea and Fusai, Niccolo and Groom, Lachy and Hausman, Karol and Ichter, Brian and others},
  journal={arXiv preprint arXiv:2410.24164},
  year={2024}
}

@article{kang2025scalable,
  title={Scalable best-of-n selection for large language models via self-certainty},
  author={Kang, Zhewei and Zhao, Xuandong and Song, Dawn},
  journal={arXiv preprint arXiv:2502.18581},
  year={2025}
}

@article{hu2024flare,
  title={Flare: Achieving masterful and adaptive robot policies with large-scale reinforcement learning fine-tuning},
  author={Hu, Jiaheng and Hendrix, Rose and Farhadi, Ali and Kembhavi, Aniruddha and Mart{\'\i}n-Mart{\'\i}n, Roberto and Stone, Peter and Zeng, Kuo-Hao and Ehsani, Kiana},
  journal={arXiv preprint arXiv:2409.16578},
  year={2024}
}

@article{lu2025vla,
  title={Vla-rl: Towards masterful and general robotic manipulation with scalable reinforcement learning},
  author={Lu, Guanxing and Guo, Wenkai and Zhang, Chubin and Zhou, Yuheng and Jiang, Haonan and Gao, Zifeng and Tang, Yansong and Wang, Ziwei},
  journal={arXiv preprint arXiv:2505.18719},
  year={2025}
}

@article{zhao2025learning,
  title={Learning to reason without external rewards},
  author={Zhao, Xuandong and Kang, Zhewei and Feng, Aosong and Levine, Sergey and Song, Dawn},
  journal={arXiv preprint arXiv:2505.19590},
  year={2025}
}

@inproceedings{cui2025grove,
  title={Grove: A generalized reward for learning open-vocabulary physical skill},
  author={Cui, Jieming and Liu, Tengyu and Meng, Ziyu and Yu, Jiale and Song, Ran and Zhang, Wei and Zhu, Yixin and Huang, Siyuan},
  booktitle={Proceedings of the Computer Vision and Pattern Recognition Conference},
  pages={15781--15790},
  year={2025}
}

@article{ma2023eureka,
  title={Eureka: Human-level reward design via coding large language models},
  author={Ma, Yecheng Jason and Liang, William and Wang, Guanzhi and Huang, De-An and Bastani, Osbert and Jayaraman, Dinesh and Zhu, Yuke and Fan, Linxi and Anandkumar, Anima},
  journal={arXiv preprint arXiv:2310.12931},
  year={2023}
}

@inproceedings{ma2024vision,
  title={Vision language models are in-context value learners},
  author={Ma, Yecheng Jason and Hejna, Joey and Fu, Chuyuan and Shah, Dhruv and Liang, Jacky and Xu, Zhuo and Kirmani, Sean and Xu, Peng and Driess, Danny and Xiao, Ted and others},
  booktitle={The Thirteenth International Conference on Learning Representations},
  year={2024}
}

@inproceedings{singh2009rewards,
  title={Where do rewards come from},
  author={Singh, Satinder and Lewis, Richard L and Barto, Andrew G},
  booktitle={Proceedings of the annual conference of the cognitive science society},
  pages={2601--2606},
  year={2009},
  organization={Cognitive Science Society}
}

@article{rocamonde2023vision,
  title={Vision-language models are zero-shot reward models for reinforcement learning},
  author={Rocamonde, Juan and Montesinos, Victoriano and Nava, Elvis and Perez, Ethan and Lindner, David},
  journal={arXiv preprint arXiv:2310.12921},
  year={2023}
}

@article{zhang2024grape,
  title={Grape: Generalizing robot policy via preference alignment},
  author={Zhang, Zijian and Zheng, Kaiyuan and Chen, Zhaorun and Jang, Joel and Li, Yi and Han, Siwei and Wang, Chaoqi and Ding, Mingyu and Fox, Dieter and Yao, Huaxiu},
  journal={arXiv preprint arXiv:2411.19309},
  year={2024}
}

@article{sutton1999policy,
  title={Policy gradient methods for reinforcement learning with function approximation},
  author={Sutton, Richard S and McAllester, David and Singh, Satinder and Mansour, Yishay},
  journal={Advances in neural information processing systems},
  volume={12},
  year={1999}
}

@inproceedings{schulman2015trust,
  title={Trust region policy optimization},
  author={Schulman, John and Levine, Sergey and Abbeel, Pieter and Jordan, Michael and Moritz, Philipp},
  booktitle={International conference on machine learning},
  pages={1889--1897},
  year={2015},
  organization={PMLR}
}

@article{schulman2017proximal,
  title={Proximal policy optimization algorithms},
  author={Schulman, John and Wolski, Filip and Dhariwal, Prafulla and Radford, Alec and Klimov, Oleg},
  journal={arXiv preprint arXiv:1707.06347},
  year={2017}
}

@inproceedings{fujimoto2018addressing,
  title={Addressing function approximation error in actor-critic methods},
  author={Fujimoto, Scott and Hoof, Herke and Meger, David},
  booktitle={International conference on machine learning},
  pages={1587--1596},
  year={2018},
  organization={PMLR}
}

@inproceedings{haarnoja2018soft,
  title={Soft actor-critic: Off-policy maximum entropy deep reinforcement learning with a stochastic actor},
  author={Haarnoja, Tuomas and Zhou, Aurick and Abbeel, Pieter and Levine, Sergey},
  booktitle={International conference on machine learning},
  pages={1861--1870},
  year={2018},
  organization={Pmlr}
}

@article{mnih2013playing,
  title={Playing atari with deep reinforcement learning},
  author={Mnih, Volodymyr and Kavukcuoglu, Koray and Silver, David and Graves, Alex and Antonoglou, Ioannis and Wierstra, Daan and Riedmiller, Martin},
  journal={arXiv preprint arXiv:1312.5602},
  year={2013}
}

@inproceedings{knox2009interactively,
  title={Interactively shaping agents via human reinforcement: The TAMER framework},
  author={Knox, W Bradley and Stone, Peter},
  booktitle={Proceedings of the fifth international conference on Knowledge capture},
  pages={9--16},
  year={2009}
}

@inproceedings{knox2012reinforcement,
  title={Reinforcement learning from simultaneous human and MDP reward.},
  author={Knox, W Bradley and Stone, Peter},
  booktitle={AAMAS},
  volume={1004},
  pages={475--482},
  year={2012},
  organization={Valencia}
}

@article{JMLR:v18:16-634,
  author  = {Christian Wirth and Riad Akrour and Gerhard Neumann and Johannes F{{\"u}}rnkranz},
  title   = {A Survey of Preference-Based Reinforcement Learning Methods},
  journal = {Journal of Machine Learning Research},
  year    = {2017},
  volume  = {18},
  number  = {136},
  pages   = {1--46},
  url     = {http://jmlr.org/papers/v18/16-634.html}
}

@article{bradley1952rank,
  title={Rank analysis of incomplete block designs: I. the method of paired comparisons},
  author={Bradley, Ralph Allan and Terry, Milton E},
  journal={Biometrika},
  volume={39},
  number={3/4},
  pages={324--345},
  year={1952},
  publisher={JSTOR}
}

@article{bai2022training,
  title={Training a helpful and harmless assistant with reinforcement learning from human feedback},
  author={Bai, Yuntao and Jones, Andy and Ndousse, Kamal and Askell, Amanda and Chen, Anna and DasSarma, Nova and Drain, Dawn and Fort, Stanislav and Ganguli, Deep and Henighan, Tom and others},
  journal={arXiv preprint arXiv:2204.05862},
  year={2022}
}

@article{rafailov2023direct,
  title={Direct preference optimization: Your language model is secretly a reward model},
  author={Rafailov, Rafael and Sharma, Archit and Mitchell, Eric and Manning, Christopher D and Ermon, Stefano and Finn, Chelsea},
  journal={Advances in neural information processing systems},
  volume={36},
  pages={53728--53741},
  year={2023}
}

@article{guo2025deepseek,
  title={Deepseek-r1: Incentivizing reasoning capability in llms via reinforcement learning},
  author={Guo, Daya and Yang, Dejian and Zhang, Haowei and Song, Junxiao and Zhang, Ruoyu and Xu, Runxin and Zhu, Qihao and Ma, Shirong and Wang, Peiyi and Bi, Xiao and others},
  journal={arXiv preprint arXiv:2501.12948},
  year={2025}
}

@article{brohan2022rt,
  title={Rt-1: Robotics transformer for real-world control at scale},
  author={Brohan, Anthony and Brown, Noah and Carbajal, Justice and Chebotar, Yevgen and Dabis, Joseph and Finn, Chelsea and Gopalakrishnan, Keerthana and Hausman, Karol and Herzog, Alex and Hsu, Jasmine and others},
  journal={arXiv preprint arXiv:2212.06817},
  year={2022}
}

@inproceedings{zitkovich2023rt,
  title={Rt-2: Vision-language-action models transfer web knowledge to robotic control},
  author={Zitkovich, Brianna and Yu, Tianhe and Xu, Sichun and Xu, Peng and Xiao, Ted and Xia, Fei and Wu, Jialin and Wohlhart, Paul and Welker, Stefan and Wahid, Ayzaan and others},
  booktitle={Conference on Robot Learning},
  pages={2165--2183},
  year={2023},
  organization={PMLR}
}

@article{huang2023embodied,
  title={An embodied generalist agent in 3d world},
  author={Huang, Jiangyong and Yong, Silong and Ma, Xiaojian and Linghu, Xiongkun and Li, Puhao and Wang, Yan and Li, Qing and Zhu, Song-Chun and Jia, Baoxiong and Huang, Siyuan},
  journal={arXiv preprint arXiv:2311.12871},
  year={2023}
}

@article{deitke2022️,
  title={ProcTHOR: Large-Scale Embodied AI Using Procedural Generation},
  author={Deitke, Matt and VanderBilt, Eli and Herrasti, Alvaro and Weihs, Luca and Ehsani, Kiana and Salvador, Jordi and Han, Winson and Kolve, Eric and Kembhavi, Aniruddha and Mottaghi, Roozbeh},
  journal={Advances in Neural Information Processing Systems},
  volume={35},
  pages={5982--5994},
  year={2022}
}

@article{richens2025general,
  title={General agents contain world models},
  author={Richens, Jonathan and Abel, David and Bellot, Alexis and Everitt, Tom},
  journal={arXiv preprint arXiv:2506.01622},
  year={2025}
}

@article{eftekhar2023selective,
  title={Selective visual representations improve convergence and generalization for embodied ai},
  author={Eftekhar, Ainaz and Zeng, Kuo-Hao and Duan, Jiafei and Farhadi, Ali and Kembhavi, Ani and Krishna, Ranjay},
  journal={arXiv preprint arXiv:2311.04193},
  year={2023}
}

@inproceedings{ramrakhya2023pirlnav,
  title={Pirlnav: Pretraining with imitation and rl finetuning for objectnav},
  author={Ramrakhya, Ram and Batra, Dhruv and Wijmans, Erik and Das, Abhishek},
  booktitle={Proceedings of the IEEE/CVF Conference on Computer Vision and Pattern Recognition},
  pages={17896--17906},
  year={2023}
}

@inproceedings{uchendu2023jump,
  title={Jump-start reinforcement learning},
  author={Uchendu, Ikechukwu and Xiao, Ted and Lu, Yao and Zhu, Banghua and Yan, Mengyuan and Simon, Jos{\'e}phine and Bennice, Matthew and Fu, Chuyuan and Ma, Cong and Jiao, Jiantao and others},
  booktitle={International Conference on Machine Learning},
  pages={34556--34583},
  year={2023},
  organization={PMLR}
}

@inproceedings{lin2024navigating,
  title={Navigating noisy feedback: Enhancing reinforcement learning with error-prone language models},
  author={Lin, Muhan and Shi, Shuyang and Guo, Yue and Chalaki, Behdad and Tadiparthi, Vaishnav and Pari, Ehsan Moradi and Stepputtis, Simon and Campbell, Joseph P and Sycara, Katia P},
  booktitle={Findings of the Association for Computational Linguistics: EMNLP 2024},
  pages={16002--16014},
  year={2024}
}

\end{document}